\documentclass[10pt,twocolumn,letterpaper]{article}

\usepackage{cvpr}              

\usepackage[dvipsnames, table]{xcolor}
\definecolor{lightgray}{gray}{0.8}
\usepackage[dvipsnames]{xcolor}

\usepackage[normalem]{ulem}
\usepackage{multirow}
\usepackage{graphicx}
\usepackage{amsmath}

\usepackage{colortbl}
\usepackage{booktabs}
\usepackage{amssymb} 
\usepackage{pifont} 
\usepackage{xcolor}
\usepackage[accsupp]{axessibility}

\definecolor{lightgray}{gray}{0.9}
\definecolor{cvprblue}{rgb}{0.21,0.49,0.74}
\usepackage[pagebackref,breaklinks,colorlinks,citecolor=cvprblue]{hyperref}

\def\paperID{35} 
\def\confName{CVPR}
\def\confYear{2024}

\title{VLM-PL: Advanced Pseudo Labeling Approach for Class Incremental Object Detection via Vision-Language Model}

\author{$^{1,2}$Junsu~kim \and $^{1,2}$Yunhoe~Ku\thanks{Equal contribution.} \and $^1$Jihyeon~Kim\footnotemark[1] \and $^1$Junuk~Cha \and $^1$Seungryul~Baek \and\and \\
{\normalsize $^1$UNIST} \qquad 
{\normalsize $^2$MODULABS} \qquad
}

\begin{document}
\maketitle
\begin{abstract}
In the field of Class Incremental Object Detection (CIOD), creating models that can continuously learn like humans is a major challenge. Pseudo-labeling methods, although initially powerful, struggle with multi-scenario incremental learning due to their tendency to forget past knowledge. To overcome this, we introduce a new approach called Vision-Language Model assisted Pseudo-Labeling (VLM-PL). This technique uses Vision-Language Model (VLM) to verify the correctness of pseudo ground-truths (GTs) without requiring additional model training. VLM-PL starts by deriving pseudo GTs from a pre-trained detector. Then, we generate custom queries for each pseudo GT using carefully designed prompt templates that combine image and text features. This allows the VLM to classify the correctness through its responses. Furthermore, VLM-PL integrates refined pseudo and real GTs from upcoming training, effectively combining new and old knowledge. Extensive experiments conducted on the Pascal VOC and MS COCO datasets not only highlight VLM-PL's exceptional performance in multi-scenario but also illuminate its effectiveness in dual-scenario by achieving state-of-the-art results in both.
\end{abstract}

\section{Introduction}
\label{sec:intro}

The pursuit of artificial intelligence that mimics human-like continuous learning has led to significant exploration in Class Incremental Learning (CIL). CIL aims to develop methods that enable models to incorporate new classes while maintaining expertise in ones already learned. This endeavor seeks to tackle a central challenge: how to expand a model's knowledge base without eroding its current knowledge, a phenomenon known as catastrophic forgetting~\cite{robins1995catastrophic}.
\begin{figure}[t]
\centering{
\includegraphics[width=1.0\linewidth]{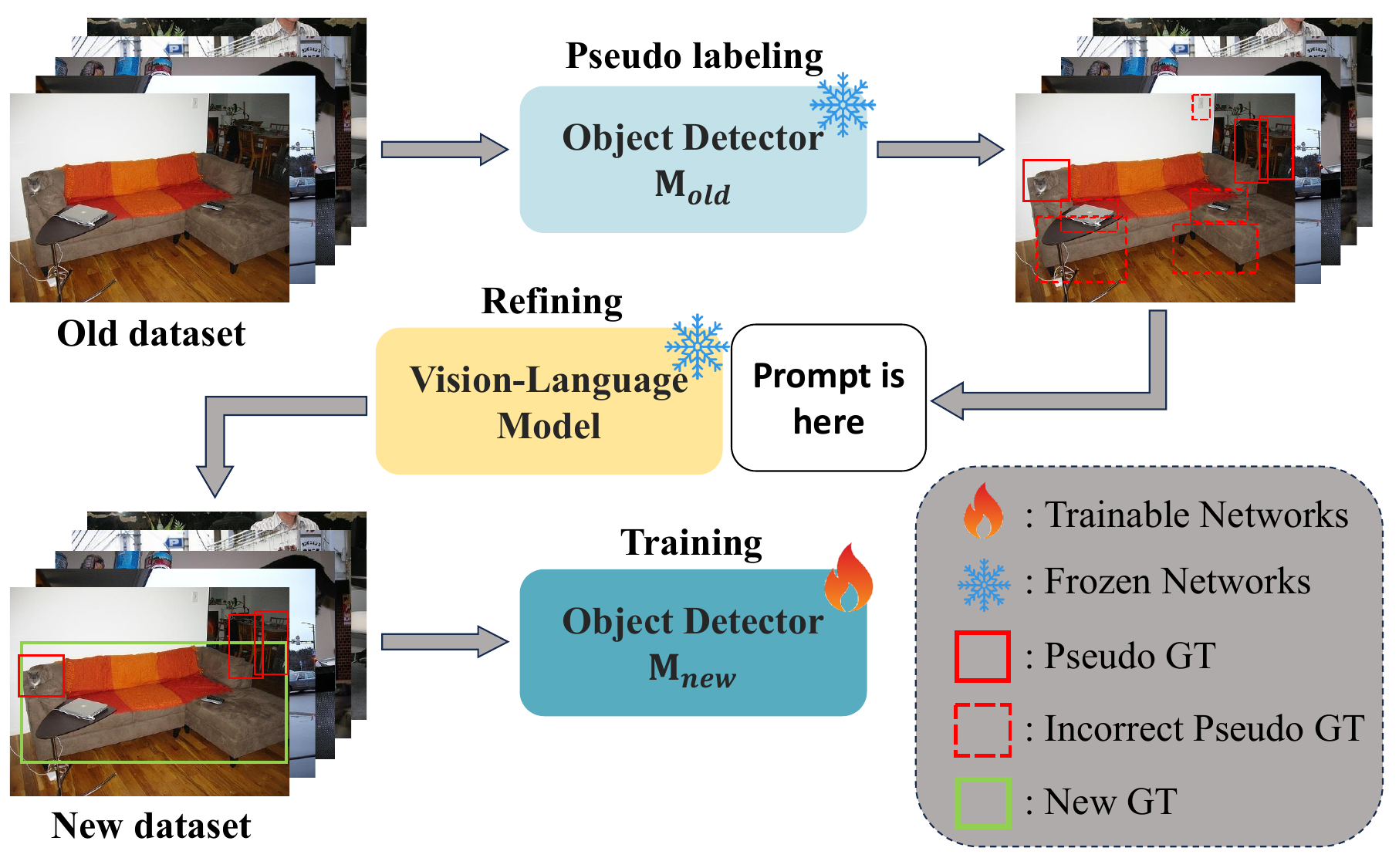}
}
    \vspace{-1em}
    \caption{\emph{Workflow of our proposed method}: This schematic illustrates the sequential steps of our method. It starts with pseudo-labeling by a pre-trained model $\mathbf{M}_{old}$, followed by refining through the Vision-Language Model. Custom-generated prompts are used for each pseudo ground-truth (GT). This refining process filters out incorrect pseudo GTs to yield reliable pseudo GTs. These annotations are then used to train a detector $\mathbf{M}_{new}$, incorporating previous knowledge with the updated dataset.}
    \vspace{-1.5em}
\label{fig:fig1-briefflow}
\end{figure}

To address this challenge, the academic community has predominantly adopted three strategies: regularization, knowledge distillation, and replay. Regularization methods~\cite{jung2016less, zenke2017synaptic, kirkpatrick2017overcoming, lopez2017gradient, paik2020overcoming, zhang2020class} aim to preserve previous learning by penalizing changes to critical parameters. Knowledge distillation~\cite{li2017learning, hao2019end, simon2021learning, lu2022augmented, rebuffi2017icarl} techniques, on the other hand, focus on transferring knowledge from an older model version to its updated form, ensuring that the new model retains the ability to perform well on old tasks. Replay approaches, categorized into partial experience replay~\cite{rebuffi2017icarl, shin2017continual, he2018exemplar, chaudhry2019continual, de2021continual} and deep generative replay~\cite{wu2018MRGAN, cong2020ganmemory, xiang2019ILCAN, shin2017dgr, gao2023ddgr, jodelet2023SDCIL}, combat catastrophic forgetting~\cite{robins1995catastrophic}. The former retains a subset of previous tasks' data, utilizing it as a memory buffer while training new tasks. The latter uses generative models to re-experience past tasks' data.

Class incremental learning (CIL) strategies, designed to embrace new knowledge while retaining previous learnings, often encounter limitations when applied to the more intricate class incremental object detection (CIOD). CIOD, which is considered a separate domain, deals with the complex challenge of detecting multiple labels within a scene. As a result, CIOD necessitates advanced methodologies to enhance detection capabilities across varied labels without compromising on the accuracy of identifying previously learned classes. 


In the field of CIOD, pioneering researches~\cite{shmelkov2017incremental,liu2020incdet, acharya2020rodeo, feng2022erders} have played a crucial role in advancing the field. These innovations extend methodologies initially crafted for single-class classification to the multifaceted challenges of CIOD, yielding notable advancements. Furthermore, the transition of detection frameworks from traditional CNN-based~\cite{li2020generalized, ren2015faster} towards transformer-based~\cite{carion2020end, zhu2020deformable, li2022dn}, introducing research directions that capitalize on the attention mechanism's ability to enhance model generalization capabilities. In this base-frameworks transition, pseudo-labeling has been utilized to boost model performance in many CIOD strategies for alleviating forgetting. For example, techniques like class-wise distillation using pseudo ground-truth (GT) in CL-DETR~\cite{liu2023CLDETR}, along with methods used in OW-DETR~\cite{gupta2022ow} and SDDGR~\cite{kim2024SDDGR}, involve pseudo labeling approach using the confidence score from the classification branch of object queries. Despite their innovative applications, these techniques are fundamentally dependent on the performance of previously trained models. This reliance introduces significant limitations, especially in multi-incremental scenarios. As the complexity of scenarios increases, the retention of knowledge from earlier tasks weakens, leading to a discernible decline in performance. This decline is attributed to inaccuracies in pseudo GTs generated based on prior models' knowledge.

To address the problems mentioned earlier, we introduce VLM-PL, a new approach method that uses Vision-Language Models~\cite{you2023ferret, liu2023llava, internlmxcomposer2} (VLMs) to enhance pseudo-labeling by refining incorrect pseudo GTs. This method ensures the consistent use of accurate pseudo ground-truths (GTs) in various scenarios. Drawing inspiration from recent studies~\cite{kim2024SDDGR, jodelet2023SDCIL, ding2022don_clip, thengane2022clip} in pre-trained foundation models~\cite{radford2021clip, rombach2022high, li2023gligen}, our approach utilizes the vast knowledge of these models. This counters the significant performance drop that pseudo-labeling strategies often face due to reliance on previously trained models, especially in multi-class incremental object detection settings. Specifically, we employ prompt-tuning with VLMs to identify reliable pseudo GTs, eliminating the need for model retraining for classification tasks. This strategy effectively reduces error accumulation in complex scenarios and exhibits strong performance in both multi and dual scenarios. Moreover, it achieves state-of-the-art results for both Pascal and COCO datasets without the need for a replay strategy. The flow of the proposed approach is illustrated in Figure~\ref{fig:fig1-briefflow}. Our contributions are organized as follows: 
\begin{itemize}
    \item To the best of our knowledge, we are pioneers in integrating VLM into CIOD, addressing challenges not primarily tackled in this field before.
    \item Our method introduces effective prompt-tuning of VLM and input-output flows that accommodate scenarios with multiple incremental class additions, thus combating the usual performance declines seen in such challenging situations.
    \item Extensive experiments show that our approach excels not only in multi-incremental scenarios but also sets a new state-of-the-art in single incremental scenarios, thereby revealing the impact of VLM assistance on object detection.
\end{itemize}

\section{Related works}
\label{sec:rw}

\subsection{Continual learning}
\label{subsec:CL}
Continual learning represents a broad spectrum of machine learning strategies designed to equip models with the capability to learn continuously, accumulating knowledge over time without catastrophic forgetting~\cite{robins1995catastrophic} the previously acquired information. Within this broad domain, class incremental learning (CIL) is a critical subset, focusing specifically on the model's ability to integrate new classes seamlessly.

\noindent{\textbf{Class incremental learning.}} 
CIL primarily addresses the challenge of classification, where models learn to identify new classes over time while retaining accuracy on previously learned classes. The essence of CIL is based on its methodologies, which can be broadly categorized into three key strategies: regularization, distillation, and replay. \emph{Regularization} methods, such as~\cite{jung2016less, zenke2017synaptic, kirkpatrick2017overcoming, lopez2017gradient, paik2020overcoming, zhang2020class}, aim to restrict the model's parameter updates to preserve knowledge across tasks. 
\emph{Distillation} methods~\cite{li2017learning, hao2019end, simon2021learning, lu2022augmented} focus on the transfer of knowledge from a teacher model to a student model, where the teacher model, which encapsulates prior knowledge, guides the student model to retain old information while learning new. \emph{Replay} involves reusing a subset of original data (partial experience replay~\cite{rebuffi2017icarl, guo2020randomsampling, bang2021rainbow, prabhu2020gdumb, koh2021blurry, chaudhry2018randomsampling}) or generating new data samples mimicking the old data distribution (generative replay~\cite{gao2023ddgr, shin2017dgr, wu2018MRGAN, cong2020ganmemory, xiang2019ILCAN}), to prevent the learned knowledge forgetting. However, expanding the principles of CIL to object detection introduces a harder challenge: class incremental object detection(CIOD). Unlike CIL, where the focus is on classifying single objects within an image, CIOD involves detecting and classifying multiple object instances across various classes within the same image. Consequently, researchers have started investigating the specific strategy for CIOD.


\noindent{\textbf{Class incremental object detection.}}
Recent developments in CIOD have explored both methodologies, such as CNN-based~\cite{chen2020ap, feng2021tood, bochkovskiy2020yolov4, lin2017feature, pang2019libra, ren2015faster, peng2020fasterILOD, cermelli2022MMA, liu2023augmented} and transformer-based~\cite{liu2023CLDETR, gupta2022ow, kang2023alleviatin, dong2022openworldDETR, kim2023classwise, kim2024SDDGR}. In general, transformer-based methods have been more focused on their superior generalization ability compared to CNN-based. Within this context, CL-DETR~\cite{liu2023CLDETR}, OW-DETR~\cite{gupta2022ow}, and Open-world DETR~\cite{dong2022openworldDETR} have effectively utilized the architectural strengths of transformers, such as pseudo-labeling and class-wise distribution replay, while kang~\etal~\cite{kang2023alleviatin} suggested both fine-grained distillations (\ie DMD and IFD) strategy. 
Despite these innovations, such methods~\cite{liu2023CLDETR, gupta2022ow, dong2022openworldDETR} often over-depend on the performance of the previous model when trained in multi incremental scenarios. To address this, we propose a new approach employing a Vision-Language Model (VLM), integrating the capabilities of a large language model and vision model, aiming to surpass the limitations inherent in reducing the dependency on the previously trained models within the pseudo-labeling strategy.
\begin{figure*}[t]
\centering{
\includegraphics[width=1.0\linewidth]{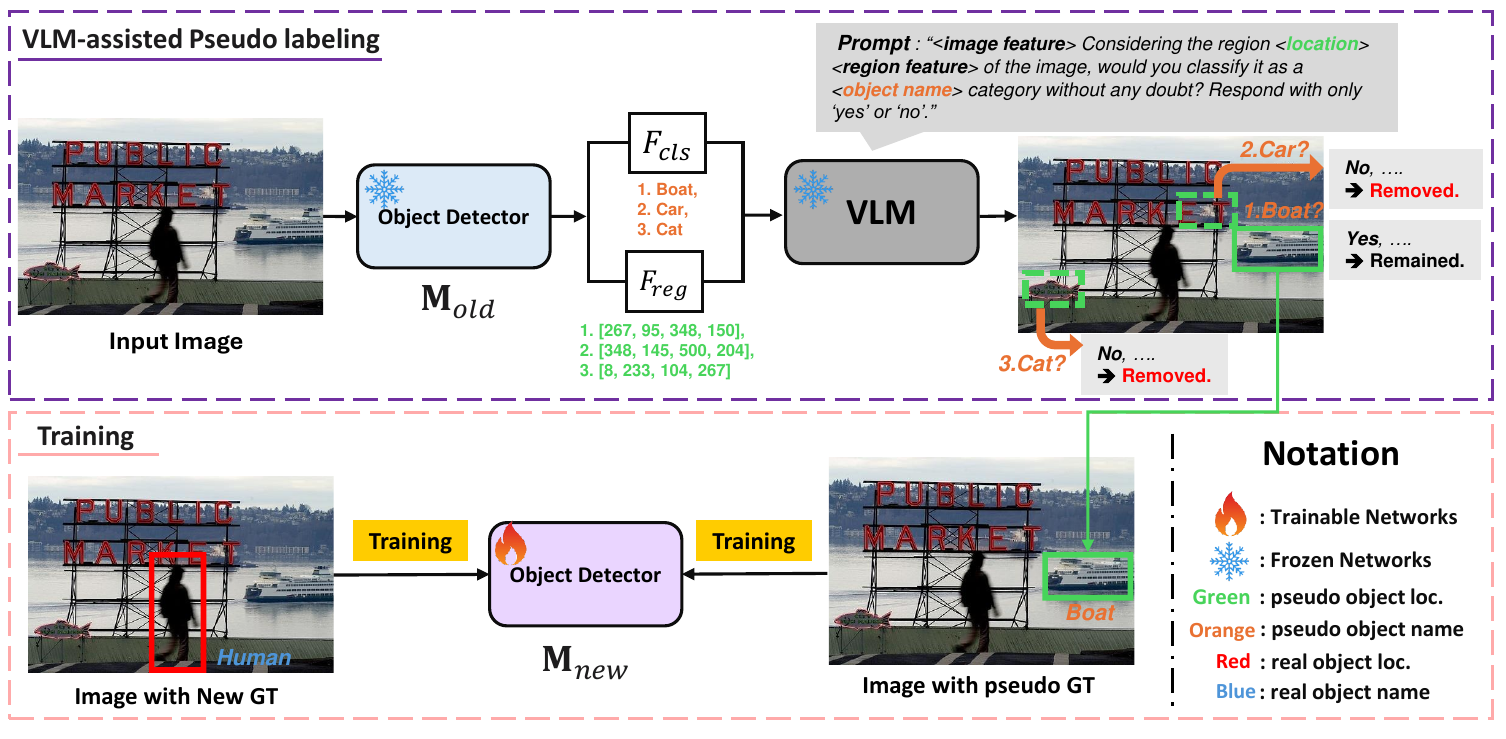}
}
    \vspace{-2em}
    \caption{\emph{Overview of the VLM-assisted Pseudo Labeling}: The sequence begins with the detector $\mathbf{M}_{old}$, applying pseudo labeling to identify potential objects (\eg \textcolor{orange}{Boat}, \textcolor{orange}{Car}, and \textcolor{orange}{Cat} within the input image), alongside their corresponding \textcolor{Green}{bounding box locations}. Each identified object and its location are encapsulated into a prompt template. This template integrates placeholders for $<$image feature$>$ and $<$region feature$>$, where the former is substituted with the overall image features and the latter with features corresponding to the specific region of interest. The prompts are classified by the VLM for reliability, using responses such as `yes' or `no' to verify each pseudo GT. Subsequently, the refined pseudo GTs are combined with new GTs from the new task for training the detector $\mathbf{M}_{new}$.}
    \vspace{-0.5em}
\label{fig:fig2-overall}
\end{figure*}

\subsection{Vision-Language Models}
In recent years, the performance of large language models (LLMs) has been advancing rapidly, leading to active research in vision models based on the superior performance of LLMs. Initially, models in the form of BLIP~\cite{blip, blip2} or GLIP~\cite{li2022grounded} were investigated, focusing on simple captioning of images. Recently, research has progressed beyond image captioning to various tasks such as question answering~\cite{liu2023llava, zhu2023minigpt, ye2023mplugowl, li2023otter, dai2023instructblip, bai2023qwenvl, vicuna2023, dong2024internlm} using CLIP~\cite{radford2021clip} and visual encoder~\cite{fang2023eva, blip, blip2}. Specifically, InternLM~\cite{dong2024internlm} achieved superior performance by proposing to train only modules that extract visual features using LoRA~\cite{hu2021lora}. Alongside, the domain of region-based question answering~\cite{peng2023kosmos, you2023ferret, chen2023shikra, you2023ferret, zhang2023gpt4roi, chen2023position, zhao2023bubogpt, wang2024visionllm, zang2023contextual} has witnessed considerable advancements, enabling detailed conversations about specific regions within an image. In particular, the Ferret~\cite{you2023ferret} proposed a spatial-aware visual sampler to enable any-prompt visual input, regardless of the format of the prompt. We consider that the large-scale knowledge inherent in model~\cite{you2023ferret} holds the potential to mitigate catastrophic forgetting, so we propose to utilize this potential with conventional method such as pseudo-labeling in our work.

\section{Preliminaries}
\subsection{Transformer-based detector}
\label{pre:DETR}
The advent of transformer-based architectures~\cite{zhang2022dino,carion2020end,zhu2020deformable, liu2022dab}, especially the Detection Transformer (DETR) series, has significantly advanced the field of object detection. DETR models utilize an efficient attention~\cite{vaswani2017attention} mechanism to diminish the inductive bias found in conventional detection frameworks~\cite{ren2015faster, redmon2018yolov3, bochkovskiy2020yolov4}, crucial for distinguishing foreground from background entities and extracting global feature relationships. Thus, it 
ultimately enhances the model's generalization capabilities~\cite{carion2020end}. Furthermore, a notable innovation is their ability to predict object categories and bounding boxes directly through self-attention and cross-attention layers, eliminating the need for traditional post-processing techniques like non-maximum suppression. This process involves processing a predefined number of object queries, each predicting class probabilities and bounding box (bbox) coordinates. As a result, DETR-based detectors simplify the pipeline and reduce the dependency on handcrafted features, ensuring more robust detection outcomes. We use Deformable DETR~\cite{zhu2020deformable} (D-DETR) for its efficient object query handling and reduced computational costs.

\subsection{Region-specific conversation}
\label{pre:VLM}
Ferret~\cite{you2023ferret} represents a significant advancement in Vision-Language Models (VLMs) by enabling precise, region-specific question-answering capabilities within images. It can detail conversation by employing a hybrid region representation, which integrates both discrete features (\eg points and boxes) and continuous features (\eg strokes, scribbles, or complex polygons). Ferret is utilizing the CLIP~\cite{radford2021clip} model for visual features, achieves a detailed capture of image features by resizing images to $336 \times 336$ and extracting feature embeddings $Z \in \mathbb{R}^{H \times W \times C}$. Here, $H$ is image height, $W$ is image width and $C$ is the image channel. Textual embeddings are derived using a pre-trained Language Model tokenizer, providing embeddings $T \in \mathbb{R}^{L \times D}$ where $L$ and $D$ denote sequence length and embedding dimension, respectively. The visual embeddings are projected to match the text embeddings dimension $D$, facilitating the seamless integration of visual and textual input. 
To integrate visual features into the language model's text prompts, Ferret employs a placeholder approach, indicated by $<$SPECIAL$>$. This allows the language model to process visual information alongside textual data. Specifically, to inform the language model of the exact location of a feature within an image, the question is formatted as ``a region $[x_1, y_1, x_2, y_2]$ $<$SPECIAL$>$", where $[x_1, y_1, x_2, y_2]$ denotes the coordinates of the region of interest. This methodology enables the model to understand and respond to queries about specific areas within an image, significantly enhancing its ability to provide precise, context-aware answers.

\section{Methods}
\subsection{Process configuration} 
\label{sec:method}
In the domain of class incremental object detection (CIOD), the primary objective is seamlessly integrating new class labels into the model without diminishing its performance on classes that have already been learned. This process is divided into a series of separate tasks, represented as $\mathcal{T}_d$, where $d$ ranges from 1 to $N$, indicating the total number of tasks. Each task, $\mathcal{T}_d$, is uniquely identified with a dataset $\mathcal{D}_d$ comprising a collection of images, denoted as $\mathcal{X}_d = \{x_{d}^1, x_{d}^2, \ldots, x_{d}^{L}\}$, and their respective annotations, $\mathcal{Y}_d = \{\mathbf{y}_{d}^1, \mathbf{y}_{d}^2, \ldots, \mathbf{y}_{d}^{L}\}$. Here, $L$ represents the number of data points for task $\mathcal{T}_d$, with $d$ serving as the index for each task. Each annotation vector $\mathbf{y}_{d}^l$ for an image $x_{d}^l$ can encapsulate multiple object instances, represented as $\mathbf{y}_{d}^l = \{(c_1, \textbf{B}_1), \ldots, (c_K, \textbf{B}_K)\}$, where $K$ denotes the number of objects in an image $l$, and each $c_k$ and $\textbf{B}_k$ denote a category and bbox coordinates of the $k$-th object, respectively. This approach is aligned with the conventional CIOD configuration as prior works~\cite{shmelkov2017incremental, feng2022erders, kang2023alleviatin, kim2024SDDGR}. 

Our VLM-PL consists of two primary components: 1) A method for extracting pseudo GTs from the pre-trained model (Section~\ref{method:pseudo labeling}), and 2) A technique for refining these pseudo GTs through the VLM model (Section~\ref{method:VLM_assistance}). A comprehensive overview can be seen in Figure~\ref{fig:fig2-overall}.

\subsection{Pseudo-labeling}
\label{method:pseudo labeling}
The adaptation of the pseudo-labeling mechanism~\cite{gupta2022ow, liu2023CLDETR, kim2024SDDGR, dong2022openworldDETR} for transformer-based detectors~\cite {hao2019end, zhu2020deformable, zhang2022dino} leverages the decoder's output from learned object queries. Each object query then goes through the classification branch ($\mathbf{F}_\text{cls}$) and regression branch ($\mathbf{F}_\text{reg}$), which include several MLPs at the final layer of the decoder. In the classification branch, queries calculate logits across all learned categories, including the background. These logits are then converted into scores ranging from 0 to 1 using a sigmoid function across all categories. This results in an output score matrix $\hat{O} \in \mathbb{R}^{Q \times (\text{category}+1)}$ for class predictions and $\hat{\mathbf{B}} \in \mathbb{R}^{Q \times 4}$ for bboxes, where $Q$ represents the total number of object queries. Building upon this framework, scores that exceed a predefined threshold, $\tau$, are nominated for pseudo-labeling, ensuring the utilization of only the most reliable predictions in generating pseudo ground-truth (pseudo GT), denoted by $\hat{\mathbf{y}} = \{\hat{c}, \hat{\mathbf{B}}\}$. Here, $\hat{c}$ represents the predicted category name (\eg `person', `kite' etc) from the highest-scoring category in each query, and $\hat{\mathbf{B}}$ provides the bbox coordinates for the corresponding query. In our study, drawing on the insights from previous research on~\cite{liu2023CLDETR, kim2024SDDGR}, we set the $\tau$ threshold to 0.3 as the optimal confidence score for maximizing performance.

\begin{figure}[t]
\centering{
\includegraphics[width=1.\linewidth]{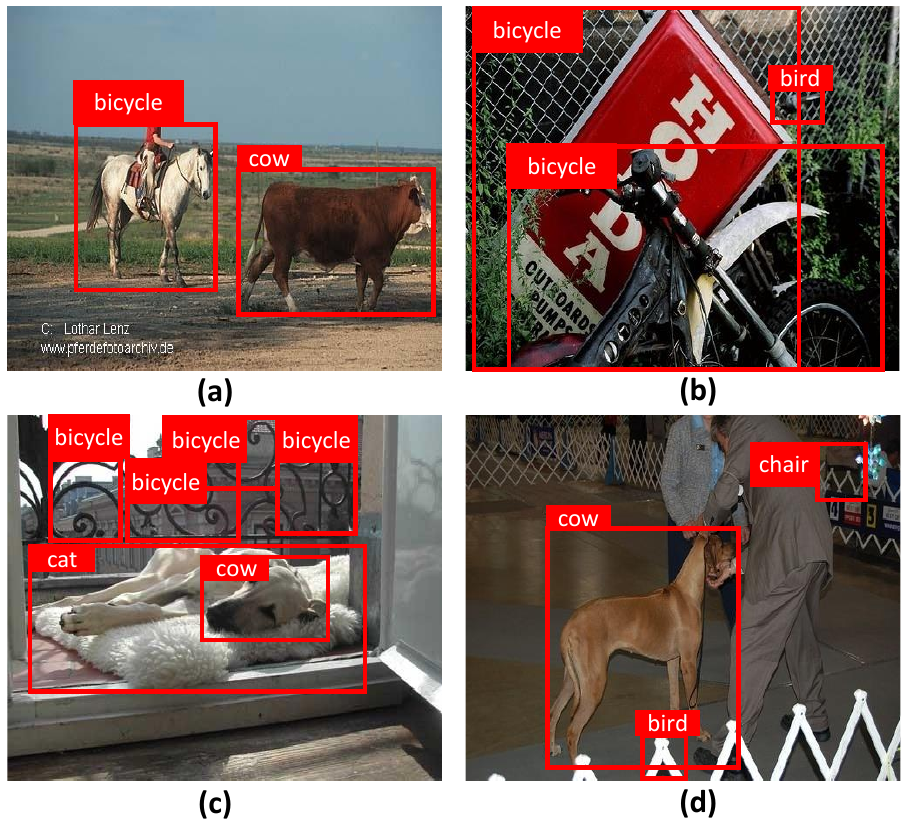}
}
    \vspace{-1em}
    \caption{Illustration of incorrect pseudo GT examples generated during a multi-incremental learning scenario (\ie 5+5+5+5) on the Pascal~\cite{everingham2010pascal} dataset. (a) depicts an incorrect pseudo GT where a `bicycle' is mislabeled; (b) shows both `bicycle' and `bird' misidentified; (c) highlights a case where all annotations are incorrect; and (d) indicates mislabeled `cow' and `bird' instances.}
\label{fig:incorrectPL}
\vspace{-1em}
\end{figure}

Following the conventional pseudo-labeling strategy~\cite{shmelkov2017incremental, liu2023CLDETR, kim2024SDDGR} for CIOD, we use the generated pseudo ground truth $\hat{y}$, in conjunction with the actual labels $y$, to train new coming categories. 
However, despite our efforts to alleviate forgetting, the reliance on predictions from previously trained models in pseudo-labeling can still potentially degrade performance. While this issue is less noticeable in dual-task scenarios, it becomes significantly more prominent when incrementally training on multiple tasks (\eg 5+5+5+5, 10+5+5 in our experiments). Because, in multi-scenarios, the knowledge from previous models about earlier learned objects becomes increasingly blurred. This leads to reliance on incorrect predictions as pseudo GT for subsequent training. This not only creates a harmful cycle of error propagation, but also exacerbates performance degradation. Figure~\ref{fig:incorrectPL} shows incorrect pseudo GT examples of conventional pseudo-labeling strategies on the Pascal~\cite{everingham2010pascal} dataset.

\begin{table*}[t]
\centering
\caption{Performance comparison of CIOD methods on Pascal VOC 2007 in \emph{multi-scenario} settings, using $AP_{50}$ (\%) metric. Results for~\cite{shmelkov2017incremental, peng2020fasterILOD, cermelli2022MMA} are cited from ABR study~\cite{liu2023augmented}. The red arrow (\textcolor{red}{\small{↑}}) indicates performance improvement over the previous state-of-the-art (SoTA). Meanwhile, a red dash (\textcolor{red}{\small{-}}) indicates performance equivalent to the SoTA. The best results in each configuration are highlighted in \textbf{bold}.}
\begin{tabular}{|c||cc|c||cc|c|}
\hline
\multicolumn{1}{|c||}{} & \multicolumn{3}{c||}{\textbf{10+5+5 (3-tasks)}} & \multicolumn{3}{c|}{\textbf{5+5+5+5 (4-tasks)}} \\
\cline{2-7}
\multicolumn{1}{|c||}{\multirow{-2}{*}{\textbf{Method}}} & \textbf{1-10 ($\mathcal{T}_1$)} & \textbf{11-20 ($\mathcal{T}_2+\mathcal{T}_3$)} & \textbf{1-20} & \textbf{1-5 ($\mathcal{T}_1$)} & \textbf{6-20 ($\mathcal{T}_2+\mathcal{T}_3+\mathcal{T}_4$)} & \textbf{1-20} \\ \hline
    ILOD~\cite{shmelkov2017incremental} & 67.2 & 59.4 & 63.3 & 58.5 & 15.6 & 26.3 \\
    Faster ILOD~\cite{peng2020fasterILOD} & 68.3 & 57.9 & 63.1 & 55.7 & 16.0 & 25.9 \\
    MMA~\cite{cermelli2022MMA} & 67.4 & 60.5 & 64.0 & 62.3 & 31.2 & 38.9 \\
    ABR~\cite{liu2023augmented} & 68.7 & 67.1 & \textbf{67.9} & 64.7 & 56.4 & 58.4 \\ 
    DMD+IFD~\cite{kang2023alleviatin} & - & - & - & 46.14 & 60.19 & 58.72 \\ 
    \rowcolor{lightgray}VLM-PL(Ours) & 67.9 & 67.9 & \textbf{67.9} \textcolor{red}{\small{-}} & 64.5 & 68.4 &\textbf{ 65.5} \textcolor{red}{\small{↑6.78}} \\ \hline
\end{tabular}
\label{tab:multiVOC}
\end{table*}


\begin{table*}[t]
\centering
\caption{Performance comparison of CIOD methods on Pascal VOC 2007 in \emph{dual-scenario} settings, utilizing the $AP_{50}$ (\%) metric.  Results for~\cite{shmelkov2017incremental, peng2020fasterILOD, cermelli2022MMA, zhou2020ppas, yang2022mvc, joseph2021ore, gupta2022ow, joseph2021metailod} are cited from the ABR study~\cite{liu2023augmented}. The red arrow (\textcolor{red}{\small{↑}}) signifies an improvement over the previous state-of-the-art. The best results in each configuration are highlighted in \textbf{bold}.}
\resizebox{2.07\columnwidth}{!}{%
\begin{tabular}{|c||cc|c||cc|c||cc|c||cc|c|}
\hline
\multicolumn{1}{|c||}{} & \multicolumn{3}{c||}{\textbf{19+1}} & \multicolumn{3}{c||}{\textbf{15+5}} & \multicolumn{3}{c||}{\textbf{10+10}} & \multicolumn{3}{c|}{\textbf{5+15}} \\
\cline{2-13}
\multicolumn{1}{|c||}{\multirow{-2}{*}{\textbf{Method}}} & \textbf{1-19} ($\mathcal{T}_1$) & \textbf{20} ($\mathcal{T}_2$) & \textbf{1-20} & \textbf{1-15} ($\mathcal{T}_1$) & \textbf{16-20} ($\mathcal{T}_2$) & \textbf{1-20} & \textbf{1-10} ($\mathcal{T}_1$) & \textbf{11-20} ($\mathcal{T}_2$) & \textbf{1-20} & \textbf{1-5} ($\mathcal{T}_1$) & \textbf{6-20} ($\mathcal{T}_2$) & \textbf{1-20} \\ \hline
    ILOD~\cite{shmelkov2017incremental} & 69.8 & 64.5 & 69.6 & 72.5 & 58.5 & 68.9 & 69.8 & 53.7 & 61.7 & 61.0 & 37.3 & 43.2 \\
    Faster ILOD~\cite{peng2020fasterILOD} & 70.9 & 63.2 & 70.6 & 73.1 & 57.3 & 69.2 & 70.3 & 53.0 & 61.7 & 62.0 & 37.1 & 43.3 \\
    PPAS~\cite{zhou2020ppas} & 70.5 & 53.0 & 69.2 & - & - & - & 63.5 & 60.0 & 61.8 & - & - & - \\
    MVC~\cite{yang2022mvc} & 70.2 & 60.6 & 69.7 & 69.4 & 57.9 & 66.5 & 66.2 & 66.0 & 66.1 & - & - & - \\
    MMA~\cite{cermelli2022MMA} & 70.9 & 62.9 & 70.5 & 72.7 & 60.6 & 69.7 & 69.8 & 63.9 & 66.8 & 66.8 & 57.2 & 59.6 \\
    ORE~\cite{joseph2021ore} & 69.4 & 60.1 & 68.9 & 71.8 & 58.7 & 68.5 & 60.4 & 68.8 & 64.6 & - & - & - \\
    OW-DETR~\cite{gupta2022ow} & 70.2 & 62.0 & 69.8 & 72.2 & 59.8 & 69.1 & 63.5 & 67.9 & 65.7 & - & - & - \\
    Meta-ILOD~\cite{joseph2021metailod} & 70.9 & 57.6 & 70.2 & 71.7 & 55.9 & 67.8 & 68.4 & 64.3 & 66.3 & - & - & - \\
    ABR~\cite{liu2023augmented} & 71.0 & 69.7 & 70.9 & 73.0 & 65.1 & 71.0 & 71.2 & 72.8 & 72.0 & 64.7 & 71.0 & 69.4 \\
    \rowcolor{lightgray}VLM-PL(Ours) & 73.7 & 89.3 & \textbf{73.6} \textcolor{red}{\small{↑2.7}} & 73.9 & 82.4 & \textbf{72.4} \textcolor{red}{\small{↑1.4}} & 80.3 & 76.3 & \textbf{78.3} \textcolor{red}{\small{↑6.3}} & 79.4 & 83.5 & \textbf{81.0} \textcolor{red}{\small{↑11.6}} \\
    \hline
\end{tabular}}
\label{tab:dualVOC}
\end{table*}

\subsection{Vision-Language Model assistance}
\label{method:VLM_assistance}
To address the aforementioned inherent challenges of pseudo-labeling in multi-task scenarios, we employ a Vision-Language Model (VLM), specifically Ferret~\cite{you2023ferret}. Ferret's ability to perform question-answering tasks on specific image regions allows for a direct assessment of pseudo GT accuracy. 
This allows us to filter for consistent and accurate pseudo GT independent of the performance of previously used models.

\noindent\textbf{Image Feature Extraction.} Initially, the CLIP~\cite{radford2021clip} model is employed to extract the overall image features, which are crucial for grasping the comprehensive context of the image. For regions requiring pseudo GT verification, we refer to the bbox coordinates, $\hat{\mathbf{B}}$, and category names, $\hat{c}$, as described in Section~\ref{method:pseudo labeling}. It is important to note that multiple instances of pseudo GT may exist within a single image. The bbox coordinates $\hat{\mathbf{B}}$, originally in the normalized format of $[x, y, w, h]$, are transformed into real-size corner coordinates $[x_1, y_1, x_2, y_2]$. This accurately outlines each region of interest, making it easier to create binary masks (\ie 0 is non-interesting region, 1 is interesting region) that emphasize the specific areas in the context of the overall image generated through CLIP~\cite{radford2021clip}.


\noindent\textbf{Prompt Formation.} We use a predefined text prompt template to construct text queries for Ferret~\cite{you2023ferret}. This template includes not only the text format but also the overall image visual feature and specific spatial information of the identified regions. We do this by using a \emph{$<$placeholder$>$} as an LLM input. The template is: \emph{``$<$image feature$>$ Considering the region $<$location$>$ $<$region feature$>$ of the image, would you classify it as a $<$object name$>$ category without any doubt? Respond with only `yes' or `no'.}" This method represents a form of prompt-tuning, a technique that enables a Large Language Model (LLM) to perform classification tasks based on the provided context without additional model training~\cite{li2021prefixprompt, shin2020autoprompt, lester2021powerprompt}. In this template, the \emph{$<$image feature$>$} placeholder represents the position of the overall image's CLIP visual feature. The \emph{$<$location$>$} placeholder gets filled with the bbox coordinates, specifically $\hat{\mathbf{B}}$ from $\hat{\mathbf{y}}$, indicating the area of interest within the image. The \emph{$<$object name$>$} corresponds to $\hat{c}$, signifying the object's category name. Finally, the \emph{$<$region feature$>$} denotes the position of features extracted from the binary image mask, highlighting the part of the overall image feature that bbox $\hat{\mathbf{B}}$ represents. The prompt-making process is iteratively done for each pseudo GTs in an image, modifying them accordingly.



\noindent\textbf{Output.} In the final stage, our method carefully selects pseudo GTs that Ferret validates with a 'yes'. We use this curated dataset to train the model on new coming classes. This approach ensures the training process uses consistently reliable pseudo GTs without the influence of a pre-trained model.


\begin{table*}[h]
\centering
\caption{Performance comparison of CIOD methods on COCO in \emph{dual-scenario} settings of 70+10, using COCO $AP$ metrics. Results for~\cite{li2017learning, li2019rilod, cermelli2022MMA, peng2021sid, feng2022erders} are cited from the CL-DETR study~\cite{liu2023CLDETR}. The red arrow (\textcolor{red}{\small{↑}}) signifies an improvement over the previous state-of-the-art. The best performance is highlighted in \textbf{bold} at each evaluation. † denotes the version without the replay strategy (\eg partial replay, synthetic replay).}
\renewcommand{\arraystretch}{1}
\small
\resizebox{0.8\textwidth}{!}{
\begin{tabular}{c|l|cccccc}
\hline\hline
{Scenarios} & \multicolumn{1}{c|}{{Method}} & $AP$ & $AP_{50}$ & $AP_{75}$ & $AP_{S}$ & $AP_{M}$ & $AP_{L}$ \\ \hline
\hline 
\multirow{9}{*}{70 + 10} 
     & LWF~\cite{li2017learning} & 7.1 & 12.4 & 7.0 & 4.8 & 9.5 & 10.0 \\ 
     & RILOD~\cite{li2019rilod} & 24.5 & 37.9 & 25.7 & 14.2 & 27.4 & 33.5 \\ 
     & MMA~\cite{cermelli2022MMA} & 30.2 & 52.1 & 31.5 & - & - & - \\ 
     & ABR~\cite{liu2023augmented} & 31.1 & 52.9 & 32.7 & - & - & - \\ 
     & SID~\cite{peng2021sid} & 32.8 & 49.0 & 35.0 & 17.1 & 36.9 & 44.5 \\ 
     & ERD~\cite{feng2022erders} & 34.9 & 51.9 & 37.4 & 18.7 & 38.8 & 45.5 \\ \cline{2-8}
     & CL-DETR\textsuperscript{†}~\cite{liu2023CLDETR} & 35.8 & 53.5 & 39.5 & 19.4 & 41.5 & 46.1 \\
     & SDDGR\textsuperscript{†}~\cite{kim2024SDDGR} & 38.6 & 56.2 & 42.1 & 22.3 & 42.1 & 50.6 \\ 
     & VLM-PL(Ours) & \textbf{39.8} \textcolor{red}{\small{↑1.2}} & \textbf{58.2} \textcolor{red}{\small{↑2.0}} & \textbf{43.3} \textcolor{red}{\small{↑1.2}} & \textbf{23.3} \textcolor{red}{\small{↑1.0}} & \textbf{43.5} \textcolor{red}{\small{↑1.4}} & \textbf{51.6} \textcolor{red}{\small{↑1.0}} \\  
     \hline \hline 

\end{tabular}}
\label{table:dualCOCO}
\end{table*}
\begin{table}[t]
\centering
\small
\caption{Ablation experiment results of different pseudo-labeling strategies on the Pascal~\cite{everingham2010pascal} dataset with \emph{multi-scenario} (\ie 5+5+5+5) are presented. The values are mean Average Precision ($AP_{50}$, \%), and we use the ferret~\cite{you2023ferret} as a VLM. The best performance is highlighted in \textbf{bold} at each evaluation.}
\label{abl:pseudo-labeling}
\resizebox{.47\textwidth}{!}{
\begin{tabular}{lcccc}
    \toprule
    \multirow{2}{*}{Method} & \multicolumn{4}{c}{Number of Classes} \\
    \cmidrule(lr){2-5}
    & 5 & 10 & 15 & 20 \\
    \midrule
    Original pseudo labeling & 82.6\% & 76.8\% & 73.6\% & 62.4\% \\
    VLM-assist pseudo labeling & 82.6\% & \textbf{79.4}\% & \textbf{77.9}\% & \textbf{65.5}\% \\
    \bottomrule

\end{tabular}}
\end{table}
\begin{table}[t]
\centering
\small
\caption{Ablation results of different VLMs~\cite{you2023ferret, internlmxcomposer2} on the Pascal~\cite{everingham2010pascal} dataset with \emph{multi-scenario} (\ie 5+5+5+5) are presented. The values are mean Average Precision ($AP_{50}$, \%). The best performance is highlighted in \textbf{bold} at each evaluation.}
\label{abl:VLM}
\begin{tabular}{lcccc}
    \toprule
    \multirow{2}{*}{Method} & \multicolumn{4}{c}{Number of Classes} \\
    \cmidrule(lr){2-5}
    & 5 & 10 & 15 & 20 \\
    \midrule
    InternLM2~\cite{internlmxcomposer2} & 82.6\% & 75.7\% & 76.5\% & 63.9\% \\
    Ferret~\cite{you2023ferret}  & 82.6\% & \textbf{79.4}\% & \textbf{77.9}\% & \textbf{65.5}\%  \\
    \bottomrule
\end{tabular}
\vspace{-1em}
\end{table}
\section{Experiments}
\label{exp}
\subsection{Dataset and metrics}
\label{sec:Dataseteval}
\noindent{\textbf{Dataset.}} Our research primarily uses the PASCAL VOC dataset~\cite{everingham2010pascal}. Known for its 20 diverse object classes, it contains 9,963 images, split into 5,011 for training and 4,952 for testing. We also use the MS COCO 2017 dataset~\cite{lin2014microsoft} as a additional experiment. The MS COCO~\cite{lin2014microsoft}, with its 80 object classes spread across 118,000 training images and 5,000 evaluation images, serves as a challenging benchmark.

\noindent{\textbf{Eval metrics.}}
From PASCAL VOC, we use the mean average precision at the IOU threshold of 0.5 ($AP_{50}$). 
From MS COCO, we also use the mean average precision at various IOU thresholds (0.5:0.95, 0.5, 0.75) and object sizes: $AP$, $AP_{50}$, $AP_{75}$, $AP_{S}$, $AP_{M}$, and $AP_{L}$. Here, $AP$ represents the mean value across all IOU thresholds from 0.5 to 0.95. 

\subsection{Implementation and experiments}
\noindent{\textbf{Implementation details.}} In our study, we employ a methodology based on Deformable-DETR~\cite{zhu2020deformable}, utilizing the pretrained ResNet-50~\cite{he2016deep} as a backbone for extracting multi-scale features. We establish the number of object queries $Q$ at 300. The AdamW~\cite{loshchilov2017decoupled} optimizer is utilized for model training, with a learning rate of 0.0002 and a weight decay parameter of 0.0001. We also set the gradient clipping parameter to 0.1. Other hyper-parameters are in line with our baseline~\cite{zhu2020deformable}. All experimental procedures are conducted using 4 NVIDIA A100 GPUs, each with a batch size of 8. For evaluation, We use a single GPU. In the VLM assistance stage, we use the CLIP-ViT-L/14~\cite{radford2021clip} image encoder to extract image feature embeddings. We also use the Ferret-13B model~\cite{you2023ferret} as a VLM, which is trained on public datasets~\cite{krishna2017VG, yu2016modelingphrase, plummer2015flickr30k} using the Vicuna~\cite{vicuna2023} LLM.

\noindent{\textbf{Scenario setup.}} 
In the \emph{multi-scenarios} setting, the process begins with initial training on a base categories set $\mathcal{T}_1$. This is followed by subsequent training phases that introduce additional categories subsets $\mathcal{T}_{2:n}$, gradually expanding the model's knowledge. For instance, scenarios like $5(\mathcal{T}_1)+5(\mathcal{T}_2)+5(\mathcal{T}_3)+5(\mathcal{T}_4)$ and $10(\mathcal{T}_1)+5(\mathcal{T}_2)+5(\mathcal{T}_3)$ are referred to as 4-task and 3-task settings in our study. In these settings, we evaluate the cumulative learning performance after $n$ phases across all integrated classes $\mathcal{T}_{1:n}$, and assess the performance on the baseline ($\mathcal{T}_1$) and the added tasks ($\mathcal{T}_2+\mathcal{T}_3$) and ($\mathcal{T}_2+\mathcal{T}_3+\mathcal{T}_4$) separately.
In the \emph{Dual-scenario} setting, training begins on subset $\mathcal{T}_1$, then introduces $\mathcal{T}_2$, represented as $\mathcal{T}_1 + \mathcal{T}_2$. Evaluation covers all knowledge like $1-20$ across all settings (\eg $19 + 1, 15 + 5, 10 + 10, 5+ 15$), including an assessment of base knowledge (\ie $19$, $15$, $10$, $5$). For COCO~\cite{lin2014microsoft}, the data configuration follows $\mathcal{T}_1+\mathcal{T}_2$, focusing evaluation solely on comprehensive model performance.

\subsection{Results and analysis}
\noindent{\textbf{Multi-scenario.}}
Table~\ref{tab:multiVOC} demonstrates the outstanding performance of our method on the PASCAL~\cite{everingham2010pascal} dataset, particularly in the complex 4-tasks scenario. Here, we achieved a 65.5\% accuracy, marking a significant 6.78\% improvement over the previous state-of-the-art (SoTA), DMD+IFD~\cite{kang2023alleviatin}. Furthermore, in the 10+5+5 scenario, we matched the ABR's~\cite{liu2023augmented} performance without the need for partial replay. These results underscore our model's ability to counteract catastrophic forgetting by leveraging VLM knowledge, thereby maintaining the robust performance of the pseudo-labeling strategy across various scenarios.

\noindent{\textbf{Dual-scenario.}}
Despite our goal of solving \emph{multi-scenario} challenges, our VLM-assisted pseudo-labeling strategy (VLM-PL) also demonstrates significant performance in single incremental task situations, as shown in Table~\ref{tab:dualVOC} and Table~\ref{table:dualCOCO}. In Table~\ref{tab:dualVOC}, compared to OW-DETR~\cite{gupta2022ow}, which uses the same baseline Deformable DETR~\cite{zhu2020deformable}, our method surpasses it in both initial class performance and overall (\ie 1-20 result) outcome on pascal~\cite{everingham2010pascal}. It also has outstanding performance against the recent SoTA, ABR~\cite{liu2023augmented}. Our performance in scenarios 19+1, 15+5, 10+10, and 5+15 (73.7\%, 73.9\%, 80.3\%, and 79.4\%, respectively) demonstrates our method's effectiveness in mitigating forgetting, even without using directly any previous data (replay-free). Furthermore, Table~\ref{table:dualCOCO} shows that our VLM-PL method outperforms various other approaches on COCO~\cite{lin2014microsoft}, including those using the same baseline~\cite{zhu2020deformable} and excluding replay strategy. Consequently, we can check the adaptability of our strategy in both dual and multi-scenario contexts.

\begin{figure*}[t]
\centering{
\includegraphics[width=1.\linewidth]{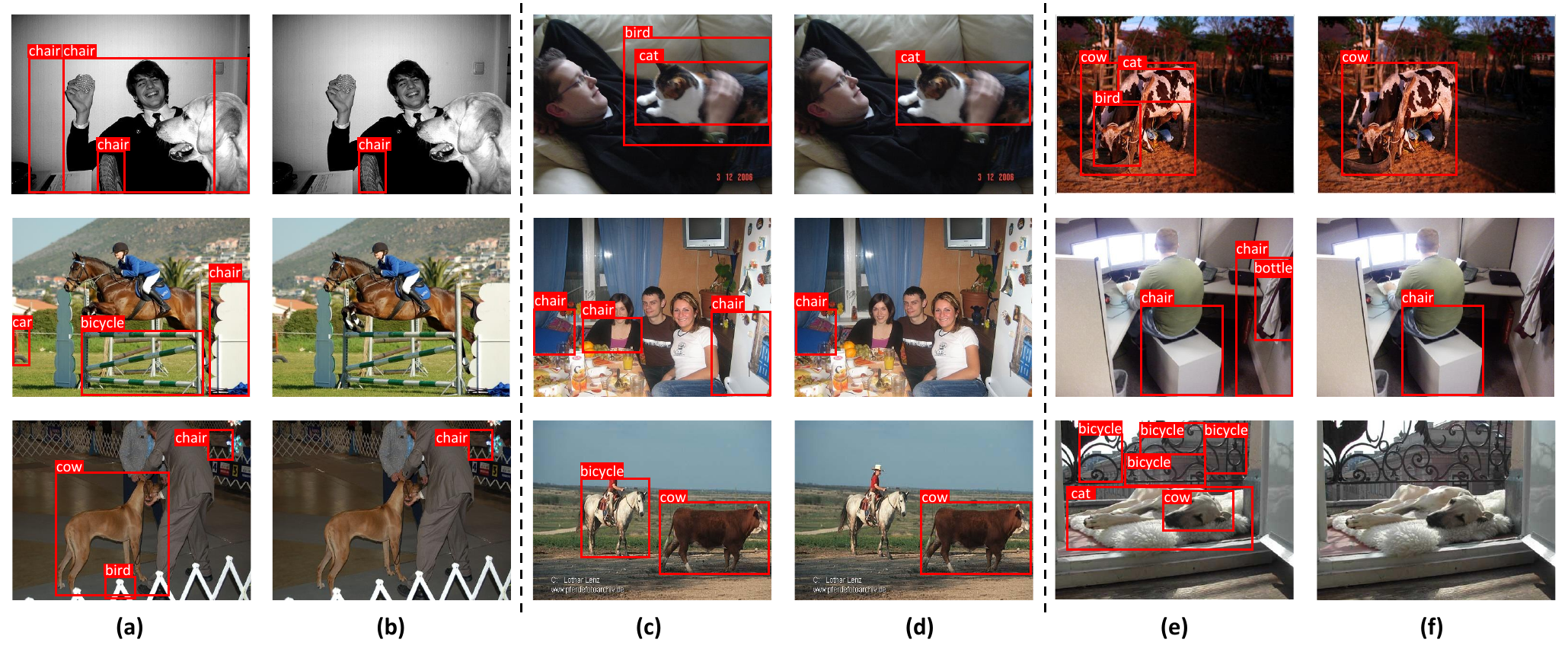}
}
    \vspace{-1.5em}
    \caption{Qualitative results of both conventional pseudo labeling, as used in OW-DETR~\cite{gupta2022ow} and SDDGR~\cite{kim2024SDDGR}, and VLM-assisted pseudo labeling in a multi-incremental scenario (for example, 5(T1)+5(T2)+5(T3)+5(T4) on the Pascal dataset) are presented here. The effects of VLM assistance can be observed from (a) to (b), (c) to (d), and (e) to (f). This is especially noticeable in (a) to (b) of the second row and (e) to (f) of the third row, which indicate that all pseudo GTs are incorrect.}
    \vspace{-1em}
\label{fig:ferretPL}
\end{figure*}
\begin{table}[t]
\centering
\small
\caption{Ablation study of the influence of VLM~\cite{you2023ferret} capacity on the Pascal~\cite{everingham2010pascal} dataset with \emph{multiple scenarios} (\ie 5+5+5+5) is presented. The values represent mean Average Precision ($AP_{50}$, \%). The best performance is highlighted in \textbf{bold} for each evaluation.}
\label{abl:capacity}
\begin{tabular}{lcccc}
    \toprule
    \multirow{2}{*}{Method} & \multicolumn{4}{c}{Number of Classes} \\
    \cmidrule(lr){2-5}
    & 5 & 10 & 15 & 20 \\
    \midrule
    Ferret 7B~\cite{you2023ferret} & 82.6\% & 78.7\% & 77.0\% & 64.7\% \\
    Ferret 13B~\cite{you2023ferret} & 82.6\% & \textbf{79.4}\% & \textbf{77.9}\% & \textbf{65.5}\% \\
    \bottomrule
\end{tabular}
\vspace{-1em}
\end{table}

\subsection{Ablations}
In our ablation study, we evaluate our components in multi-scenario settings. Specifically, we use a 4-tasks scenario on the Pascal with the metric $AP_{50}$. It's important to note that each result represents the cumulative performance up to the evaluated task (\eg T3 model evaluation includes T1, T2, and T3 classes).

\noindent{\textbf{Pseudo-labeling.}}
In Table~\ref{abl:pseudo-labeling}, we evaluate the effectiveness of using our VLM-PL approach, as compared to conventional pseudo-labeling methods. We observed that with the progression of tasks, the application of the Ferret model for refining pseudo GTs led to improvements in performance about all each tasks, demonstrating gains of 3.4\%, 4.3\%, and 3.1\% for T2, T3, and T4 respectively. These enhancements indicate that as tasks accumulate, there is a tendency for the generation of incorrect pseudo GTs, likely due to a blurring of the knowledge initially trained into the model. Nevertheless, the VLM~\cite{you2023ferret} has consistently demonstrated its ability to correct these inaccuracies, emphasizing its crucial role in refining the pseudo-labeling strategy.

\noindent{\textbf{Comparative analysis of VLMs.}}
VLMs employ various approaches, one of which is the Ferret~\cite{you2023ferret} model. This model allows detailed conversation about a specific location based on the whole image context. As indicated in Table~\ref{abl:VLM}, we compared the Ferret model with another VLM model, InternLM2~\cite{internlmxcomposer2}. It has delivered state-of-the-art results in question-answering tasks using entire images only. To use InternLM2~\cite{internlmxcomposer2} for classifying a specific image region, we cropped the image based on bboxes, identical to Ferret's approach in ours. Consequently, in the incremental tasks T2, T3, and T4, Ferret~\cite{you2023ferret} outperforms InternLM2~\cite{internlmxcomposer2} by 3.7\%, 1.4\%, and 1.6\% respectively. This indicates Ferret's effectiveness, which allows querying both the entire image feature and specific parts.
\vspace{-0.5em}

\noindent{\textbf{VLM capacity impact.}}
VLM generally relies on the performance of the LLM. Generally, LLM's larger capacities and more comprehensive training data lead to better outcomes in VLM. We investigate the impact of model capacity on the refining process of pseudo GTs. Table~\ref{abl:capacity} shows the results from experiments using Ferret~\cite{you2023ferret} 7B and Ferret~\cite{you2023ferret} 13B, each using Vicuna~\cite{vicuna2023} 7B and Vicuna~\cite{vicuna2023} 13B, respectively. Importantly, the larger 13B model provides modest but consistent improvements across all tasks, with performance increases of 0.7\%, 0.9\%, and 0.8\%. These improvements mean the LLM capacity can affect VLM performance. As a result, we chose the 13B model for our study.
\vspace{-0.5em}

\noindent{\textbf{Qualitative results.}}
As illustrated in Figure~\ref{fig:incorrectPL}, the advantage of our method is particularly noticeable in multi-scenario settings, where incorrect pseudo GTs are prevalent. The images in columns (b), (d), and (f) clearly demonstrate the thorough removal of all incorrect pseudo GTs through VLM assitance. Furthermore, in row 1, images a, c, and e vividly depict this challenge, showcasing multiple pseudo GTs for a single object. Nevertheless, through the VLM-assisted refining process, our approach successfully consolidates these into a single accurate GT for each object, demonstrating the precision and reliability of VLM assistance in the pseudo GT refining process.
\section{Conclusions And Limitations}
In this study, we introduced VLM-PL, a new approach that leverages Vision-Language Model to address the rapid performance decline in pseudo-labeling strategy under the influence of previously trained models in multiple incrementally class incremental object detection. Our VLM-PL not only demonstrates robust performance across multi and dual scenarios but also achieves state-of-the-art results in conditions without a replay strategy.

\noindent{\textbf{Limitations.}} While our approach effectively addresses pseudo-labeling reliability in various scenarios, it does have limitations. Its performance may be limited when there are few images available for pseudo-labeling, or when new images contain few objects from previously learned categories. Despite these challenges, our approach significantly outperforms not only techniques that do not utilize pseudo-labeling but also surpasses prior works~\cite{liu2023CLDETR, kim2024SDDGR} that generate pseudo GT leveraging trained model knowledge. Furthermore, we believe it also has the potential to improve methods of preventing catastrophic forgetting when used in conjunction with partial replay techniques.

\section{Acknowledgements.} This research was supported by Brian Impact Foundation, a non-profit organization dedicated to the advancement of science and technology for all. This work was supported by IITP grants (No. 2020-0-01336 Artificial intelligence graduate school program (UNIST)) and received support from LG Electronics.
\newpage
{
    \small
    \bibliographystyle{ieeenat_fullname}
    \bibliography{main}

\begin{thebibliography}{97}
\providecommand{\natexlab}[1]{#1}
\providecommand{\url}[1]{\texttt{#1}}
\expandafter\ifx\csname urlstyle\endcsname\relax
  \providecommand{\doi}[1]{doi: #1}\else
  \providecommand{\doi}{doi: \begingroup \urlstyle{rm}\Url}\fi

\bibitem[Acharya et~al.(2020)Acharya, Hayes, and Kanan]{acharya2020rodeo}
Manoj Acharya, Tyler~L Hayes, and Christopher Kanan.
\newblock Rodeo: Replay for online object detection.
\newblock In \emph{BMVC}, 2020.

\bibitem[Bai et~al.(2023)Bai, Bai, Yang, Wang, Tan, Wang, Lin, Zhou, and
  Zhou]{bai2023qwenvl}
Jinze Bai, Shuai Bai, Shusheng Yang, Shijie Wang, Sinan Tan, Peng Wang, Junyang
  Lin, Chang Zhou, and Jingren Zhou.
\newblock Qwen-vl: A versatile vision-language model for understanding,
  localization, text reading, and beyond, 2023.

\bibitem[Bang et~al.(2021)Bang, Kim, Yoo, Ha, and Choi]{bang2021rainbow}
Jihwan Bang, Heesu Kim, YoungJoon Yoo, Jung-Woo Ha, and Jonghyun Choi.
\newblock Rainbow memory: Continual learning with a memory of diverse samples.
\newblock In \emph{CVPR}, 2021.

\bibitem[Bochkovskiy et~al.(2020)Bochkovskiy, Wang, and
  Liao]{bochkovskiy2020yolov4}
Alexey Bochkovskiy, Chien-Yao Wang, and Hong-Yuan~Mark Liao.
\newblock Yolov4: Optimal speed and accuracy of object detection.
\newblock In \emph{arXiv}, 2020.

\bibitem[Carion et~al.(2020)Carion, Massa, Synnaeve, Usunier, Kirillov, and
  Zagoruyko]{carion2020end}
Nicolas Carion, Francisco Massa, Gabriel Synnaeve, Nicolas Usunier, Alexander
  Kirillov, and Sergey Zagoruyko.
\newblock End-to-end object detection with transformers.
\newblock In \emph{ECCV}, 2020.

\bibitem[Cermelli et~al.(2022)Cermelli, Geraci, Fontanel, and
  Caputo]{cermelli2022MMA}
Fabio Cermelli, Antonino Geraci, Dario Fontanel, and Barbara Caputo.
\newblock Modeling missing annotations for incremental learning in object
  detection.
\newblock In \emph{CVPRW}, 2022.

\bibitem[Chaudhry et~al.(2018)Chaudhry, Dokania, Ajanthan, and
  Torr]{chaudhry2018randomsampling}
Arslan Chaudhry, Puneet~K Dokania, Thalaiyasingam Ajanthan, and Philip~HS Torr.
\newblock Riemannian walk for incremental learning: Understanding forgetting
  and intransigence.
\newblock In \emph{ECCV}, 2018.

\bibitem[Chaudhry et~al.(2019)Chaudhry, Rohrbach, Elhoseiny, Ajanthan, Dokania,
  Torr, and Ranzato]{chaudhry2019continual}
Arslan Chaudhry, Marcus Rohrbach, Mohamed Elhoseiny, Thalaiyasingam Ajanthan,
  Puneet~K Dokania, Philip~HS Torr, and M Ranzato.
\newblock Continual learning with tiny episodic memories.
\newblock In \emph{arXiv}, 2019.

\bibitem[Chen et~al.(2023{\natexlab{a}})Chen, Qin, Luo, Mi, Li, Sun, and
  Liu]{chen2023position}
Chi Chen, Ruoyu Qin, Fuwen Luo, Xiaoyue Mi, Peng Li, Maosong Sun, and Yang Liu.
\newblock Position-enhanced visual instruction tuning for multimodal large
  language models.
\newblock \emph{arXiv preprint arXiv:2308.13437}, 2023{\natexlab{a}}.

\bibitem[Chen et~al.(2020)Chen, Lin, Li, See, Wang, and Zou]{chen2020ap}
Kean Chen, Weiyao Lin, Jianguo Li, John See, Ji Wang, and Junni Zou.
\newblock Ap-loss for accurate one-stage object detection.
\newblock In \emph{TPAMI}, 2020.

\bibitem[Chen et~al.(2023{\natexlab{b}})Chen, Zhang, Zeng, Zhang, Zhu, and
  Zhao]{chen2023shikra}
Keqin Chen, Zhao Zhang, Weili Zeng, Richong Zhang, Feng Zhu, and Rui Zhao.
\newblock Shikra: Unleashing multimodal llm's referential dialogue magic.
\newblock \emph{arXiv preprint arXiv:2306.15195}, 2023{\natexlab{b}}.

\bibitem[Chiang et~al.(2023)Chiang, Li, Lin, Sheng, Wu, Zhang, Zheng, Zhuang,
  Zhuang, Gonzalez, Stoica, and Xing]{vicuna2023}
Wei-Lin Chiang, Zhuohan Li, Zi Lin, Ying Sheng, Zhanghao Wu, Hao Zhang, Lianmin
  Zheng, Siyuan Zhuang, Yonghao Zhuang, Joseph~E. Gonzalez, Ion Stoica, and
  Eric~P. Xing.
\newblock Vicuna: An open-source chatbot impressing gpt-4 with 90\%* chatgpt
  quality, 2023.

\bibitem[Cong et~al.(2020)Cong, Zhao, Li, Wang, and Carin]{cong2020ganmemory}
Yulai Cong, Miaoyun Zhao, Jianqiao Li, Sijia Wang, and Lawrence Carin.
\newblock Gan memory with no forgetting.
\newblock In \emph{NeurIPS}, 2020.

\bibitem[Dai et~al.(2023)Dai, Li, Li, Tiong, Zhao, Wang, Li, Fung, and
  Hoi]{dai2023instructblip}
Wenliang Dai, Junnan Li, Dongxu Li, Anthony Meng~Huat Tiong, Junqi Zhao,
  Weisheng Wang, Boyang Li, Pascale Fung, and Steven Hoi.
\newblock Instructblip: Towards general-purpose vision-language models with
  instruction tuning, 2023.

\bibitem[De~Lange and Tuytelaars(2021)]{de2021continual}
Matthias De~Lange and Tinne Tuytelaars.
\newblock Continual prototype evolution: Learning online from non-stationary
  data streams.
\newblock In \emph{ICCV}, 2021.

\bibitem[Ding et~al.(2022)Ding, Liu, Tian, Yang, and Ding]{ding2022don_clip}
Yuxuan Ding, Lingqiao Liu, Chunna Tian, Jingyuan Yang, and Haoxuan Ding.
\newblock Don't stop learning: Towards continual learning for the clip model.
\newblock In \emph{arXiv}, 2022.

\bibitem[Dong et~al.(2022)Dong, Zhang, Ding, and Lee]{dong2022openworldDETR}
Na Dong, Yongqiang Zhang, Mingli Ding, and Gim~Hee Lee.
\newblock Open world detr: transformer based open world object detection.
\newblock In \emph{arXiv}, 2022.

\bibitem[Dong et~al.(2024{\natexlab{a}})Dong, Zhang, Zang, Cao, Wang, Ouyang,
  Wei, Zhang, Duan, Cao, Zhang, Li, Yan, Gao, Zhang, Li, Li, Chen, He, Zhang,
  Qiao, Lin, and Wang]{internlmxcomposer2}
Xiaoyi Dong, Pan Zhang, Yuhang Zang, Yuhang Cao, Bin Wang, Linke Ouyang, Xilin
  Wei, Songyang Zhang, Haodong Duan, Maosong Cao, Wenwei Zhang, Yining Li, Hang
  Yan, Yang Gao, Xinyue Zhang, Wei Li, Jingwen Li, Kai Chen, Conghui He,
  Xingcheng Zhang, Yu Qiao, Dahua Lin, and Jiaqi Wang.
\newblock Internlm-xcomposer2: Mastering free-form text-image composition and
  comprehension in vision-language large model.
\newblock In \emph{arXiv}, 2024{\natexlab{a}}.

\bibitem[Dong et~al.(2024{\natexlab{b}})Dong, Zhang, Zang, Cao, Wang, Ouyang,
  Wei, Zhang, Duan, Cao, et~al.]{dong2024internlm}
Xiaoyi Dong, Pan Zhang, Yuhang Zang, Yuhang Cao, Bin Wang, Linke Ouyang, Xilin
  Wei, Songyang Zhang, Haodong Duan, Maosong Cao, et~al.
\newblock Internlm-xcomposer2: Mastering free-form text-image composition and
  comprehension in vision-language large model.
\newblock \emph{arXiv preprint arXiv:2401.16420}, 2024{\natexlab{b}}.

\bibitem[Everingham et~al.(2010)Everingham, Van~Gool, Williams, Winn, and
  Zisserman]{everingham2010pascal}
Mark Everingham, Luc Van~Gool, Christopher~KI Williams, John Winn, and Andrew
  Zisserman.
\newblock The pascal visual object classes (voc) challenge.
\newblock In \emph{IJCV}, 2010.

\bibitem[Fang et~al.(2023)Fang, Wang, Xie, Sun, Wu, Wang, Huang, Wang, and
  Cao]{fang2023eva}
Yuxin Fang, Wen Wang, Binhui Xie, Quan Sun, Ledell Wu, Xinggang Wang, Tiejun
  Huang, Xinlong Wang, and Yue Cao.
\newblock Eva: Exploring the limits of masked visual representation learning at
  scale.
\newblock In \emph{CVPR}, 2023.

\bibitem[Feng et~al.(2021)Feng, Zhong, Gao, Scott, and Huang]{feng2021tood}
Chengjian Feng, Yujie Zhong, Yu Gao, Matthew~R Scott, and Weilin Huang.
\newblock Tood: Task-aligned one-stage object detection.
\newblock In \emph{ICCV}, 2021.

\bibitem[Feng et~al.(2022)Feng, Wang, and Yuan]{feng2022erders}
Tao Feng, Mang Wang, and Hangjie Yuan.
\newblock Overcoming catastrophic forgetting in incremental object detection
  via elastic response distillation.
\newblock In \emph{CVPR}, 2022.

\bibitem[Gao and Liu(2023)]{gao2023ddgr}
Rui Gao and Weiwei Liu.
\newblock Ddgr: Continual learning with deep diffusion-based generative replay.
\newblock In \emph{ICML}, 2023.

\bibitem[Guo et~al.(2020)Guo, Liu, Yang, and Rosing]{guo2020randomsampling}
Yunhui Guo, Mingrui Liu, Tianbao Yang, and Tajana Rosing.
\newblock Improved schemes for episodic memory-based lifelong learning.
\newblock In \emph{NeurIPS}, 2020.

\bibitem[Gupta et~al.(2022)Gupta, Narayan, Joseph, Khan, Khan, and
  Shah]{gupta2022ow}
Akshita Gupta, Sanath Narayan, KJ Joseph, Salman Khan, Fahad~Shahbaz Khan, and
  Mubarak Shah.
\newblock Ow-detr: Open-world detection transformer.
\newblock In \emph{CVPR}, 2022.

\bibitem[Hao et~al.(2019)Hao, Fu, Jiang, and Tian]{hao2019end}
Yu Hao, Yanwei Fu, Yu-Gang Jiang, and Qi Tian.
\newblock An end-to-end architecture for class-incremental object detection
  with knowledge distillation.
\newblock In \emph{ICME}, 2019.

\bibitem[He et~al.(2018)He, Wang, Shan, and Chen]{he2018exemplar}
Chen He, Ruiping Wang, Shiguang Shan, and Xilin Chen.
\newblock Exemplar-supported generative reproduction for class incremental
  learning.
\newblock In \emph{BMVC}, 2018.

\bibitem[He et~al.(2016)He, Zhang, Ren, and Sun]{he2016deep}
Kaiming He, Xiangyu Zhang, Shaoqing Ren, and Jian Sun.
\newblock Deep residual learning for image recognition.
\newblock In \emph{CVPR}, 2016.

\bibitem[Hu et~al.(2021)Hu, Shen, Wallis, Allen-Zhu, Li, Wang, Wang, and
  Chen]{hu2021lora}
Edward~J Hu, Yelong Shen, Phillip Wallis, Zeyuan Allen-Zhu, Yuanzhi Li, Shean
  Wang, Lu Wang, and Weizhu Chen.
\newblock Lora: Low-rank adaptation of large language models.
\newblock \emph{arXiv preprint arXiv:2106.09685}, 2021.

\bibitem[Jodelet et~al.(2023)Jodelet, Liu, Phua, and Murata]{jodelet2023SDCIL}
Quentin Jodelet, Xin Liu, Yin~Jun Phua, and Tsuyoshi Murata.
\newblock Class-incremental learning using diffusion model for distillation and
  replay.
\newblock In \emph{ICCVW}, 2023.

\bibitem[Joseph et~al.(2021{\natexlab{a}})Joseph, Khan, Khan, and
  Balasubramanian]{joseph2021ore}
KJ Joseph, Salman Khan, Fahad~Shahbaz Khan, and Vineeth~N Balasubramanian.
\newblock Towards open world object detection.
\newblock In \emph{CVPR}, 2021{\natexlab{a}}.

\bibitem[Joseph et~al.(2021{\natexlab{b}})Joseph, Rajasegaran, Khan, Khan, and
  Balasubramanian]{joseph2021metailod}
KJ Joseph, Jathushan Rajasegaran, Salman Khan, Fahad~Shahbaz Khan, and
  Vineeth~N Balasubramanian.
\newblock Incremental object detection via meta-learning.
\newblock In \emph{TPAMI}, 2021{\natexlab{b}}.

\bibitem[Jung et~al.(2016)Jung, Ju, Jung, and Kim]{jung2016less}
Heechul Jung, Jeongwoo Ju, Minju Jung, and Junmo Kim.
\newblock Less-forgetting learning in deep neural networks.
\newblock In \emph{arXiv}, 2016.

\bibitem[Kang et~al.(2023)Kang, Zhang, Zhang, Wang, Chen, Ma, and
  Huang]{kang2023alleviatin}
Mengxue Kang, Jinpeng Zhang, Jinming Zhang, Xiashuang Wang, Yang Chen, Zhe Ma,
  and Xuhui Huang.
\newblock Alleviating catastrophic forgetting of incremental object detection
  via within-class and between-class knowledge distillation.
\newblock In \emph{ICCV}, 2023.

\bibitem[Kim et~al.(2024{\natexlab{a}})Kim, Cho, Kim, Tiruneh, and
  Baek]{kim2024SDDGR}
Junsu Kim, Hoseong Cho, Jihyeon Kim, Yihalem~Yimolal Tiruneh, and Seungryul
  Baek.
\newblock Sddgr: Stable diffusion-based deep generative replay for class
  incremental object detection.
\newblock In \emph{arXiv}, 2024{\natexlab{a}}.

\bibitem[Kim et~al.(2024{\natexlab{b}})Kim, Hong, Kim, Kim, Tiruneh, On, Song,
  Choi, and Baek]{kim2023classwise}
Junsu Kim, Sumin Hong, Chanwoo Kim, Jihyeon Kim, Yihalem~Yimolal Tiruneh,
  Jeongwan On, Jihyun Song, Sunhwa Choi, and Seungryul Baek.
\newblock Class-wise buffer management for incremental object detection: An
  effective buffer training strategy.
\newblock In \emph{ICASSP}, 2024{\natexlab{b}}.

\bibitem[Kirkpatrick et~al.(2017)Kirkpatrick, Pascanu, Rabinowitz, Veness,
  Desjardins, Rusu, Milan, Quan, Ramalho, Grabska-Barwinska,
  et~al.]{kirkpatrick2017overcoming}
James Kirkpatrick, Razvan Pascanu, Neil Rabinowitz, Joel Veness, Guillaume
  Desjardins, Andrei~A Rusu, Kieran Milan, John Quan, Tiago Ramalho, Agnieszka
  Grabska-Barwinska, et~al.
\newblock Overcoming catastrophic forgetting in neural networks.
\newblock In \emph{PNAS}, 2017.

\bibitem[Koh et~al.(2021)Koh, Kim, Ha, and Choi]{koh2021blurry}
Hyunseo Koh, Dahyun Kim, Jung-Woo Ha, and Jonghyun Choi.
\newblock Online continual learning on class incremental blurry task
  configuration with anytime inference.
\newblock In \emph{arXiv}, 2021.

\bibitem[Krishna et~al.(2017)Krishna, Zhu, Groth, Johnson, Hata, Kravitz, Chen,
  Kalantidis, Li, Shamma, et~al.]{krishna2017VG}
Ranjay Krishna, Yuke Zhu, Oliver Groth, Justin Johnson, Kenji Hata, Joshua
  Kravitz, Stephanie Chen, Yannis Kalantidis, Li-Jia Li, David~A Shamma, et~al.
\newblock Visual genome: Connecting language and vision using crowdsourced
  dense image annotations.
\newblock In \emph{IJCV}, 2017.

\bibitem[Lester et~al.(2021)Lester, Al-Rfou, and
  Constant]{lester2021powerprompt}
Brian Lester, Rami Al-Rfou, and Noah Constant.
\newblock The power of scale for parameter-efficient prompt tuning.
\newblock \emph{arXiv}, 2021.

\bibitem[Li et~al.(2023{\natexlab{a}})Li, Zhang, Chen, Wang, Yang, and
  Liu]{li2023otter}
Bo Li, Yuanhan Zhang, Liangyu Chen, Jinghao Wang, Jingkang Yang, and Ziwei Liu.
\newblock Otter: A multi-modal model with in-context instruction tuning,
  2023{\natexlab{a}}.

\bibitem[Li et~al.(2019)Li, Tasci, Ghosh, Zhu, Zhang, and Heck]{li2019rilod}
Dawei Li, Serafettin Tasci, Shalini Ghosh, Jingwen Zhu, Junting Zhang, and
  Larry Heck.
\newblock Rilod: Near real-time incremental learning for object detection at
  the edge.
\newblock In \emph{SEC}, 2019.

\bibitem[Li et~al.(2022{\natexlab{a}})Li, Zhang, Liu, Guo, Ni, and
  Zhang]{li2022dn}
Feng Li, Hao Zhang, Shilong Liu, Jian Guo, Lionel~M Ni, and Lei Zhang.
\newblock Dn-detr: Accelerate detr training by introducing query denoising.
\newblock In \emph{CVPR}, 2022{\natexlab{a}}.

\bibitem[Li et~al.(2022{\natexlab{b}})Li, Li, Xiong, and Hoi]{blip}
Junnan Li, Dongxu Li, Caiming Xiong, and Steven Hoi.
\newblock Blip: Bootstrapping language-image pre-training for unified
  vision-language understanding and generation.
\newblock In \emph{ICML}, 2022{\natexlab{b}}.

\bibitem[Li et~al.(2023{\natexlab{b}})Li, Li, Savarese, and Hoi]{blip2}
Junnan Li, Dongxu Li, Silvio Savarese, and Steven Hoi.
\newblock Blip-2: Bootstrapping language-image pre-training with frozen image
  encoders and large language models.
\newblock In \emph{arXiv}, 2023{\natexlab{b}}.

\bibitem[Li et~al.(2022{\natexlab{c}})Li, Zhang, Zhang, Yang, Li, Zhong, Wang,
  Yuan, Zhang, Hwang, et~al.]{li2022grounded}
Liunian~Harold Li, Pengchuan Zhang, Haotian Zhang, Jianwei Yang, Chunyuan Li,
  Yiwu Zhong, Lijuan Wang, Lu Yuan, Lei Zhang, Jenq-Neng Hwang, et~al.
\newblock Grounded language-image pre-training.
\newblock In \emph{CVPR}, 2022{\natexlab{c}}.

\bibitem[Li et~al.(2020)Li, Wang, Wu, Chen, Hu, Li, Tang, and
  Yang]{li2020generalized}
Xiang Li, Wenhai Wang, Lijun Wu, Shuo Chen, Xiaolin Hu, Jun Li, Jinhui Tang,
  and Jian Yang.
\newblock Generalized focal loss: Learning qualified and distributed bounding
  boxes for dense object detection.
\newblock In \emph{NeurIPS}, 2020.

\bibitem[Li and Liang(2021)]{li2021prefixprompt}
Xiang~Lisa Li and Percy Liang.
\newblock Prefix-tuning: Optimizing continuous prompts for generation.
\newblock \emph{arXiv}, 2021.

\bibitem[Li et~al.(2023{\natexlab{c}})Li, Liu, Wu, Mu, Yang, Gao, Li, and
  Lee]{li2023gligen}
Yuheng Li, Haotian Liu, Qingyang Wu, Fangzhou Mu, Jianwei Yang, Jianfeng Gao,
  Chunyuan Li, and Yong~Jae Lee.
\newblock Gligen: Open-set grounded text-to-image generation.
\newblock In \emph{CVPR}, 2023{\natexlab{c}}.

\bibitem[Li and Hoiem(2017)]{li2017learning}
Zhizhong Li and Derek Hoiem.
\newblock Learning without forgetting.
\newblock In \emph{TPAMI}, 2017.

\bibitem[Lin et~al.(2014)Lin, Maire, Belongie, Hays, Perona, Ramanan,
  Doll{\'a}r, and Zitnick]{lin2014microsoft}
Tsung-Yi Lin, Michael Maire, Serge Belongie, James Hays, Pietro Perona, Deva
  Ramanan, Piotr Doll{\'a}r, and C~Lawrence Zitnick.
\newblock Microsoft coco: Common objects in context.
\newblock In \emph{ECCV}, 2014.

\bibitem[Lin et~al.(2017)Lin, Doll{\'a}r, Girshick, He, Hariharan, and
  Belongie]{lin2017feature}
Tsung-Yi Lin, Piotr Doll{\'a}r, Ross Girshick, Kaiming He, Bharath Hariharan,
  and Serge Belongie.
\newblock Feature pyramid networks for object detection.
\newblock In \emph{CVPR}, 2017.

\bibitem[Liu et~al.(2023{\natexlab{a}})Liu, Li, Wu, and Lee]{liu2023llava}
Haotian Liu, Chunyuan Li, Qingyang Wu, and Yong~Jae Lee.
\newblock Visual instruction tuning.
\newblock In \emph{NeurIPS}, 2023{\natexlab{a}}.

\bibitem[Liu et~al.(2020)Liu, Kuang, Chen, Xue, Yang, and Zhang]{liu2020incdet}
Liyang Liu, Zhanghui Kuang, Yimin Chen, Jing-Hao Xue, Wenming Yang, and Wayne
  Zhang.
\newblock Incdet: In defense of elastic weight consolidation for incremental
  object detection.
\newblock In \emph{TNNLS}, 2020.

\bibitem[Liu et~al.(2022)Liu, Li, Zhang, Yang, Qi, Su, Zhu, and
  Zhang]{liu2022dab}
Shilong Liu, Feng Li, Hao Zhang, Xiao Yang, Xianbiao Qi, Hang Su, Jun Zhu, and
  Lei Zhang.
\newblock Dab-detr: Dynamic anchor boxes are better queries for detr.
\newblock In \emph{ICLR}, 2022.

\bibitem[Liu et~al.(2023{\natexlab{b}})Liu, Cong, Goswami, Liu, and van~de
  Weijer]{liu2023augmented}
Yuyang Liu, Yang Cong, Dipam Goswami, Xialei Liu, and Joost van~de Weijer.
\newblock Augmented box replay: Overcoming foreground shift for incremental
  object detection.
\newblock In \emph{ICCV}, 2023{\natexlab{b}}.

\bibitem[Liu et~al.(2023{\natexlab{c}})Liu, Schiele, Vedaldi, and
  Rupprecht]{liu2023CLDETR}
Yaoyao Liu, Bernt Schiele, Andrea Vedaldi, and Christian Rupprecht.
\newblock Continual detection transformer for incremental object detection.
\newblock In \emph{CVPR}, 2023{\natexlab{c}}.

\bibitem[Lopez-Paz and Ranzato(2017)]{lopez2017gradient}
David Lopez-Paz and Marc'Aurelio Ranzato.
\newblock Gradient episodic memory for continual learning.
\newblock In \emph{NeurIPS}, 2017.

\bibitem[Loshchilov and Hutter(2017)]{loshchilov2017decoupled}
Ilya Loshchilov and Frank Hutter.
\newblock Decoupled weight decay regularization.
\newblock In \emph{arXiv}, 2017.

\bibitem[Lu et~al.(2022)Lu, Wang, and Deng]{lu2022augmented}
Yichen Lu, Mei Wang, and Weihong Deng.
\newblock Augmented geometric distillation for data-free incremental person
  reid.
\newblock In \emph{CVPR}, 2022.

\bibitem[Paik et~al.(2020)Paik, Oh, Kwak, and Kim]{paik2020overcoming}
Inyoung Paik, Sangjun Oh, Taeyeong Kwak, and Injung Kim.
\newblock Overcoming catastrophic forgetting by neuron-level plasticity
  control.
\newblock In \emph{AAAI}, 2020.

\bibitem[Pang et~al.(2019)Pang, Chen, Shi, Feng, Ouyang, and
  Lin]{pang2019libra}
Jiangmiao Pang, Kai Chen, Jianping Shi, Huajun Feng, Wanli Ouyang, and Dahua
  Lin.
\newblock Libra r-cnn: Towards balanced learning for object detection.
\newblock In \emph{CVPR}, 2019.

\bibitem[Peng et~al.(2020)Peng, Zhao, and Lovell]{peng2020fasterILOD}
Can Peng, Kun Zhao, and Brian~C Lovell.
\newblock Faster ilod: Incremental learning for object detectors based on
  faster rcnn.
\newblock In \emph{Pattern Recognition}, 2020.

\bibitem[Peng et~al.(2021)Peng, Zhao, Maksoud, Li, and Lovell]{peng2021sid}
Can Peng, Kun Zhao, Sam Maksoud, Meng Li, and Brian~C Lovell.
\newblock Sid: Incremental learning for anchor-free object detection via
  selective and inter-related distillation.
\newblock In \emph{CVIU}, 2021.

\bibitem[Peng et~al.(2023)Peng, Wang, Dong, Hao, Huang, Ma, and
  Wei]{peng2023kosmos}
Zhiliang Peng, Wenhui Wang, Li Dong, Yaru Hao, Shaohan Huang, Shuming Ma, and
  Furu Wei.
\newblock Kosmos-2: Grounding multimodal large language models to the world.
\newblock \emph{arXiv preprint arXiv:2306.14824}, 2023.

\bibitem[Plummer et~al.(2015)Plummer, Wang, Cervantes, Caicedo, Hockenmaier,
  and Lazebnik]{plummer2015flickr30k}
Bryan~A Plummer, Liwei Wang, Chris~M Cervantes, Juan~C Caicedo, Julia
  Hockenmaier, and Svetlana Lazebnik.
\newblock Flickr30k entities: Collecting region-to-phrase correspondences for
  richer image-to-sentence models.
\newblock In \emph{ICCV}, 2015.

\bibitem[Prabhu et~al.(2020)Prabhu, Torr, and Dokania]{prabhu2020gdumb}
Ameya Prabhu, Philip~HS Torr, and Puneet~K Dokania.
\newblock Gdumb: A simple approach that questions our progress in continual
  learning.
\newblock In \emph{ECCV}, 2020.

\bibitem[Radford et~al.(2021)Radford, Kim, Hallacy, Ramesh, Goh, Agarwal,
  Sastry, Askell, Mishkin, Clark, et~al.]{radford2021clip}
Alec Radford, Jong~Wook Kim, Chris Hallacy, Aditya Ramesh, Gabriel Goh,
  Sandhini Agarwal, Girish Sastry, Amanda Askell, Pamela Mishkin, Jack Clark,
  et~al.
\newblock Learning transferable visual models from natural language
  supervision.
\newblock In \emph{ICML}, 2021.

\bibitem[Rebuffi et~al.(2017)Rebuffi, Kolesnikov, Sperl, and
  Lampert]{rebuffi2017icarl}
Sylvestre-Alvise Rebuffi, Alexander Kolesnikov, Georg Sperl, and Christoph~H
  Lampert.
\newblock icarl: Incremental classifier and representation learning.
\newblock In \emph{CVPR}, 2017.

\bibitem[Redmon and Farhadi(2018)]{redmon2018yolov3}
Joseph Redmon and Ali Farhadi.
\newblock Yolov3: An incremental improvement.
\newblock In \emph{arXiv}, 2018.

\bibitem[Ren et~al.(2015)Ren, He, Girshick, and Sun]{ren2015faster}
Shaoqing Ren, Kaiming He, Ross Girshick, and Jian Sun.
\newblock Faster r-cnn: Towards real-time object detection with region proposal
  networks.
\newblock In \emph{NeurIPS}, 2015.

\bibitem[Robins(1995)]{robins1995catastrophic}
Anthony Robins.
\newblock Catastrophic forgetting, rehearsal and pseudorehearsal.
\newblock In \emph{Connection Science}, 1995.

\bibitem[Rombach et~al.(2022)Rombach, Blattmann, Lorenz, Esser, and
  Ommer]{rombach2022high}
Robin Rombach, Andreas Blattmann, Dominik Lorenz, Patrick Esser, and Bj{\"o}rn
  Ommer.
\newblock High-resolution image synthesis with latent diffusion models.
\newblock In \emph{CVPR}, 2022.

\bibitem[Shin et~al.(2017{\natexlab{a}})Shin, Lee, Kim, and
  Kim]{shin2017continual}
Hanul Shin, Jung~Kwon Lee, Jaehong Kim, and Jiwon Kim.
\newblock Continual learning with deep generative replay.
\newblock In \emph{NeurIPS}, 2017{\natexlab{a}}.

\bibitem[Shin et~al.(2017{\natexlab{b}})Shin, Lee, Kim, and Kim]{shin2017dgr}
Hanul Shin, Jung~Kwon Lee, Jaehong Kim, and Jiwon Kim.
\newblock Continual learning with deep generative replay.
\newblock In \emph{NeurIPS}, 2017{\natexlab{b}}.

\bibitem[Shin et~al.(2020)Shin, Razeghi, Logan~IV, Wallace, and
  Singh]{shin2020autoprompt}
Taylor Shin, Yasaman Razeghi, Robert~L Logan~IV, Eric Wallace, and Sameer
  Singh.
\newblock Autoprompt: Eliciting knowledge from language models with
  automatically generated prompts.
\newblock \emph{arXiv}, 2020.

\bibitem[Shmelkov et~al.(2017)Shmelkov, Schmid, and
  Alahari]{shmelkov2017incremental}
Konstantin Shmelkov, Cordelia Schmid, and Karteek Alahari.
\newblock Incremental learning of object detectors without catastrophic
  forgetting.
\newblock In \emph{ICCV}, 2017.

\bibitem[Simon et~al.(2021)Simon, Koniusz, and Harandi]{simon2021learning}
Christian Simon, Piotr Koniusz, and Mehrtash Harandi.
\newblock On learning the geodesic path for incremental learning.
\newblock In \emph{CVPR}, 2021.

\bibitem[Thengane et~al.(2022)Thengane, Khan, Hayat, and
  Khan]{thengane2022clip}
Vishal Thengane, Salman Khan, Munawar Hayat, and Fahad Khan.
\newblock Clip model is an efficient continual learner.
\newblock In \emph{arXiv}, 2022.

\bibitem[Vaswani et~al.(2017)Vaswani, Shazeer, Parmar, Uszkoreit, Jones, Gomez,
  Kaiser, and Polosukhin]{vaswani2017attention}
Ashish Vaswani, Noam Shazeer, Niki Parmar, Jakob Uszkoreit, Llion Jones,
  Aidan~N Gomez, {\L}ukasz Kaiser, and Illia Polosukhin.
\newblock Attention is all you need.
\newblock In \emph{NeurIPS}, 2017.

\bibitem[Wang et~al.(2024)Wang, Chen, Chen, Wu, Zhu, Zeng, Luo, Lu, Zhou, Qiao,
  et~al.]{wang2024visionllm}
Wenhai Wang, Zhe Chen, Xiaokang Chen, Jiannan Wu, Xizhou Zhu, Gang Zeng, Ping
  Luo, Tong Lu, Jie Zhou, Yu Qiao, et~al.
\newblock Visionllm: Large language model is also an open-ended decoder for
  vision-centric tasks.
\newblock \emph{Advances in Neural Information Processing Systems}, 36, 2024.

\bibitem[Wu et~al.(2018)Wu, Herranz, Liu, Van De~Weijer, Raducanu,
  et~al.]{wu2018MRGAN}
Chenshen Wu, Luis Herranz, Xialei Liu, Joost Van De~Weijer, Bogdan Raducanu,
  et~al.
\newblock Memory replay gans: Learning to generate new categories without
  forgetting.
\newblock In \emph{NeurIPS}, 2018.

\bibitem[Xiang et~al.(2019)Xiang, Fu, Ji, and Huang]{xiang2019ILCAN}
Ye Xiang, Ying Fu, Pan Ji, and Hua Huang.
\newblock Incremental learning using conditional adversarial networks.
\newblock In \emph{NeurIPS}, 2019.

\bibitem[Yang et~al.(2022)Yang, Zhou, Zhang, Sun, Wu, Wang, and
  Ye]{yang2022mvc}
Dongbao Yang, Yu Zhou, Aoting Zhang, Xurui Sun, Dayan Wu, Weiping Wang, and
  Qixiang Ye.
\newblock Multi-view correlation distillation for incremental object detection.
\newblock In \emph{Pattern Recognition}, 2022.

\bibitem[Ye et~al.(2023)Ye, Xu, Xu, Ye, Yan, Zhou, Wang, Hu, Shi, Shi, Li, Xu,
  Chen, Tian, Qi, Zhang, and Huang]{ye2023mplugowl}
Qinghao Ye, Haiyang Xu, Guohai Xu, Jiabo Ye, Ming Yan, Yiyang Zhou, Junyang
  Wang, Anwen Hu, Pengcheng Shi, Yaya Shi, Chenliang Li, Yuanhong Xu, Hehong
  Chen, Junfeng Tian, Qian Qi, Ji Zhang, and Fei Huang.
\newblock mplug-owl: Modularization empowers large language models with
  multimodality, 2023.

\bibitem[You et~al.(2024)You, Zhang, Gan, Du, Zhang, Wang, Cao, Chang, and
  Yang]{you2023ferret}
Haoxuan You, Haotian Zhang, Zhe Gan, Xianzhi Du, Bowen Zhang, Zirui Wang,
  Liangliang Cao, Shih-Fu Chang, and Yinfei Yang.
\newblock Ferret: Refer and ground anything anywhere at any granularity.
\newblock In \emph{ICLR}, 2024.

\bibitem[Yu et~al.(2016)Yu, Poirson, Yang, Berg, and
  Berg]{yu2016modelingphrase}
Licheng Yu, Patrick Poirson, Shan Yang, Alexander~C Berg, and Tamara~L Berg.
\newblock Modeling context in referring expressions.
\newblock In \emph{ECCV}, 2016.

\bibitem[Zang et~al.(2023)Zang, Li, Han, Zhou, and Loy]{zang2023contextual}
Yuhang Zang, Wei Li, Jun Han, Kaiyang Zhou, and Chen~Change Loy.
\newblock Contextual object detection with multimodal large language models.
\newblock \emph{arXiv preprint arXiv:2305.18279}, 2023.

\bibitem[Zenke et~al.(2017)Zenke, Poole, and Ganguli]{zenke2017synaptic}
Friedemann Zenke, Ben Poole, and Surya Ganguli.
\newblock Continual learning through synaptic intelligence.
\newblock In \emph{ICML}, 2017.

\bibitem[Zhang et~al.(2022)Zhang, Li, Liu, Zhang, Su, Zhu, Ni, and
  Shum]{zhang2022dino}
Hao Zhang, Feng Li, Shilong Liu, Lei Zhang, Hang Su, Jun Zhu, Lionel Ni, and
  Harry Shum.
\newblock Dino: Detr with improved denoising anchor boxes for end-to-end object
  detection.
\newblock In \emph{ICLR}, 2022.

\bibitem[Zhang et~al.(2020)Zhang, Zhang, Ghosh, Li, Tasci, Heck, Zhang, and
  Kuo]{zhang2020class}
Junting Zhang, Jie Zhang, Shalini Ghosh, Dawei Li, Serafettin Tasci, Larry
  Heck, Heming Zhang, and C-C~Jay Kuo.
\newblock Class-incremental learning via deep model consolidation.
\newblock In \emph{WACV}, 2020.

\bibitem[Zhang et~al.(2023)Zhang, Sun, Chen, Xiao, Shao, Zhang, Chen, and
  Luo]{zhang2023gpt4roi}
Shilong Zhang, Peize Sun, Shoufa Chen, Min Xiao, Wenqi Shao, Wenwei Zhang, Kai
  Chen, and Ping Luo.
\newblock Gpt4roi: Instruction tuning large language model on
  region-of-interest.
\newblock \emph{arXiv preprint arXiv:2307.03601}, 2023.

\bibitem[Zhao et~al.(2023)Zhao, Lin, Zhou, Huang, Feng, and
  Kang]{zhao2023bubogpt}
Yang Zhao, Zhijie Lin, Daquan Zhou, Zilong Huang, Jiashi Feng, and Bingyi Kang.
\newblock Bubogpt: Enabling visual grounding in multi-modal llms.
\newblock \emph{arXiv preprint arXiv:2307.08581}, 2023.

\bibitem[Zhou et~al.(2020)Zhou, Chang, Sosa, Hamann, and Cox]{zhou2020ppas}
Wang Zhou, Shiyu Chang, Norma Sosa, Hendrik Hamann, and David Cox.
\newblock Lifelong object detection.
\newblock In \emph{arXiv}, 2020.

\bibitem[Zhu et~al.(2023)Zhu, Chen, Shen, Li, and Elhoseiny]{zhu2023minigpt}
Deyao Zhu, Jun Chen, Xiaoqian Shen, Xiang Li, and Mohamed Elhoseiny.
\newblock Minigpt-4: Enhancing vision-language understanding with advanced
  large language models.
\newblock In \emph{arXiv}, 2023.

\bibitem[Zhu et~al.(2020)Zhu, Su, Lu, Li, Wang, and Dai]{zhu2020deformable}
Xizhou Zhu, Weijie Su, Lewei Lu, Bin Li, Xiaogang Wang, and Jifeng Dai.
\newblock Deformable detr: Deformable transformers for end-to-end object
  detection.
\newblock In \emph{ICLR}, 2020.

\end{thebibliography}
}

\end{document}